# Katyusha X: Practical Momentum Method for Stochastic Sum-of-Nonconvex Optimization

(version 1)

Zeyuan Allen-Zhu
zeyuan@csail.mit.edu
Microsoft Research AI

February 9, 2018

**Abstract**

The problem of minimizing sum-of-nonconvex functions (i.e., convex functions that are average of non-convex ones) is becoming increasingly important in machine learning, and is the core machinery for PCA, SVD, regularized Newton's method, accelerated non-convex optimization, and more.

We show how to provably obtain an accelerated stochastic algorithm for minimizing sum-of-nonconvex functions, by *adding one additional line* to the well-known SVRG method. This line corresponds to momentum, and shows how to directly apply momentum to the finite-sum stochastic minimization of sum-of-nonconvex functions. As a side result, our method enjoys linear parallel speed-up using mini-batch.

## 1 Introduction

The diverse world of non-convex machine learning tasks have given rise to numerous non-convex optimization problems. Some of them are perhaps as hard as minimizing general non-convex objectives (such as deep learning), but some others may be only slightly harder than convex optimization (such as matrix completion, principal component analysis, dictionary learning, etc). Therefore, it is both interesting and challenging to identify classes of optimization problems that *interplay* between non-convex and convex optimization, and (hopefully) optimally and practically solving them.

At least tracing back to 2015, Shalev-Shwartz [27] identified a class of functions that are convex, but can be written as finite average of non-convex functions. That is,[1]

$$\min_{x \in \mathbb{R}^d} \big\{ f(x) = \frac{1}{n} \sum_{i=1}^{n} f_i(x) \big\} \tag{1.1}$$

where each $f_i(x)$ is smooth and non-convex, but their average $f(x) = \frac{1}{n}\sum_{i=1}^n f_i(x)$ is convex.

We say $f(x)$ a <u>sum-of-nonconvex</u> (but convex) function following [11, 14].

In this paper, we show how to provably obtain an *accelerated* and *stochastic* method for minimizing (1.1). Our new method is based on adding only *one* additional line to the well-known SVRG (stochastic variance-reduction gradient) method [16, 32]. This additional line corresponds to momentum. To the best of our knowledge, this explains for the first time how to directly apply momentum to the stochastic minimization of sum-of-nonconvex functions. We hope this new algorithm and the new insight of this paper could facilitate our understanding towards how to correctly and provably apply momentum to non-convex machine learning tasks.

---

[1]In fact, we study a more general composite minimization setting $\min_{x \in \mathbb{R}^d} \big\{ F(x) \stackrel{\text{def}}{=} \psi(x) + \frac{1}{n}\sum_{i \in [n]} f_i(x) \big\}$ where $\psi(x)$ is some proper convex function. In this high-level introduction, we ignore the $\psi(\cdot)$ term for simplicity.

## 1.1 Motivating Examples

There is an increasing number of machine learning tasks that are found reducible to minimizing sum-of-nonconvex functions. For most such tasks, the only known approach for achieving accelerated stochastic performance relies on solving Problem (1.1).

Perhaps the most famous example is the *shift-and-invert* approach to solve PCA [26]. Let $A = \frac{1}{n} \sum_{i=1}^n a_i a_i^\top \in \mathbb{R}^{d \times d}$ be some covariance matrix and $\lambda$ be its largest eigenvalue. Then, computing $A$'s top eigenvector reduces to applying power method to a new matrix $B = (\mu I - A)^{-1}$ with $\mu = \lambda_{\max}(A) \cdot (1 + \delta)$ for some approximation parameter $\delta > 0$ [14]. In other words, PCA reduces to repeatedly minimizing convex functions $f(x) \stackrel{\text{def}}{=} \frac{1}{2} x^\top (\mu I - A) x + b^\top x$ for different vectors $b$. If one defines $f_i(x) \stackrel{\text{def}}{=} \frac{1}{2} x^\top (\mu I - a_i a_i^\top) x + b^\top x$, then $f_i(x)$ is smooth and non-convex, but $f(x)$ is convex.

- HIGH ACCURACY NEEDED. In the above reduction, we need very high accuracy (e.g., $10^{-10}$) when minimizing $f(x)$, because we need to apply power method on $B$ so the error blows up. This is very different from classical empirical minimization problems (such as Lasso, SVM), where we only need to minimize the training objective to some accuracy such as $10^{-3}$.
- NECESSITY OF PROBLEM (1.1) While there are many algorithms to solve PCA, to the best of our knowledge, the only known stochastic method which gives a provable accelerated rate requires solving Problem (1.1). We discuss more in Related Works.

Other problems that reduce to Problem (1.1) include:

- The accelerated stochastic algorithms for computing top $k$ principle components ($k$-PCA) and top $k$ singular vectors ($k$-SVD) require solving Problem (1.1) [7].
- The fastest way to compute the near-optimal strategy for the online eigenvector problem (against an adversarial opponent) is by solving Problem (1.1) [9].
- Up to this date, the fastest finite-sum stochastic algorithm for finding approximate local minima of a general non-convex smooth function is either based on cubic regularized Newton's method [1] or a special reduction [12]. Both approaches require solving Problem (1.1).
- In certain parameter regimes, the fastest finite-sum stochastic algorithm for minimizing "approximately convex functions" is based on solving Problem (1.1) [3, 12].

## 1.2 Known Approaches

In the online stochastic setting (i.e., when $n$ is infinite), there is hardly any difference between $f_i(x)$ being convex or not. The stochastic gradient descent (SGD) method gives a $T \propto \varepsilon^{-2}$ convergence rate to Problem (1.1), or a $T \propto (\sigma \varepsilon)^{-1}$ rate if $f(x)$ is $\sigma$-strongly convex. Both rates are optimal, regardless of $f_i(x)$ being convex or not. Even in the case of finding a point with small gradient, there is no difference between $f_i(x)$ being convex or not [4].

**Variance Reduction.** In the finite-sum stochastic setting (i.e., when $n$ is finite), it was discovered by Shalev-Shwartz [27] and Garber et al. [14] that one can solve (1.1) using variance reduction: for instance, using the SVRG method that was originally designed for convex optimization [16, 32].

Specifically, if $f(x)$ is $\sigma$-strongly convex and each $f_i(x)$ is $L$-smooth, then the SVRG method converges to an $\varepsilon$-minimizer of Problem (1.1) using $T_{\mathsf{grad}}$ stochastic gradient computations, where

$$T_{\mathsf{grad}} = O\Big(\big(n + \frac{\sqrt{n}L}{\sigma}\big) \log \frac{1}{\varepsilon}\Big) \text{ if } \sigma > 0 \quad \text{or} \quad T_{\mathsf{grad}} = O\Big(n \log \frac{1}{\varepsilon} + \frac{\sqrt{n}L}{\varepsilon}\Big) \text{ if } \sigma = 0 \quad . \quad (1.2)$$

Both rates outperform their corresponding counterparts in the SGD case.

*Remark* 1.1. The two complexity bounds in (1.2) do not seem to be recorded anywhere. In particular, Shalev-Shwartz [27] and Garber et al. [14] showed $T_{\mathsf{grad}} = O\big((n+\frac{L^2}{\sigma^2})\log\frac{1}{\varepsilon}\big)$ in the case of $\sigma > 0$; and Allen-Zhu and Yuan [11] showed $T_{\mathsf{grad}} = O\big(n \log \frac{1}{\varepsilon} + \frac{L^2}{\varepsilon^2}\big)$ in the case of $\sigma = 0$. Both complexity bounds are worse than (1.2). We prove (1.2) as a side result of this paper in Appendix D.

**How to Accelerate.** If Nesterov's accelerated gradient method [22, 23] (also known as the *momentum* method) is used, one can achieve $T_{\mathsf{grad}} = O\big(\frac{n\sqrt{L}}{\sqrt{\sigma}} \log \frac{1}{\varepsilon}\big)$ and $T_{\mathsf{grad}} = O\big(\frac{n\sqrt{L}}{\sqrt{\varepsilon}}\big)$ in the two cases. This square-root dependence on $\sigma$ or $\varepsilon$ is known as the *accelerated convergence rate*.

However, Nesterov's method is not stochastic and its $T_{\mathsf{grad}}$ linearly scales with $n$. Can we design a stochastic first-order method that also has an accelerated convergence rate?

*Remark* 1.2. Obtaining accelerated rates is crucial for Problem (1.1), because high accuracy is usually needed as we discussed in Section 1.1.

This can be partially answered by the APPA [13] and Catalyst [18] reduction, that we both refer to as Catalyst.[2] Mathematically, Catalyst turns any non-accelerated method into an accelerated one. For instance, when $\sigma > 0$, Catalyst on SVRG (often referred to as AccSVRG) uses the following logic:

— Define a "new problem" $\arg\min_x \{f(x) + \frac{L'}{2}\|x - \widehat{x}\|^2\}$ for some $\widehat{x} \in \mathbb{R}^d$ and $L' = L/\sqrt{n}$.

— Use Nesterov's method to minimize $f(x)$, which requires solving the "new problem" $\widetilde{O}\big(1+\frac{\sqrt{L'}}{\sqrt{\sigma}}\big)$ times.

— Solve each "new problem" by SVRG: (1.2) gives $T'_{\mathsf{grad}} = \widetilde{O}\big(\big(n + \frac{\sqrt{n}L}{L'}\big)\big) = \widetilde{O}(n)$.

In total, this requires $T_{\mathsf{grad}}$ stochastic gradient computations for $T_{\mathsf{grad}} = \widetilde{O}\big(\big(1 + \frac{\sqrt{L'}}{\sqrt{\sigma}}\big) \times T'_{\mathsf{grad}}\big)$. In other words, AccSVRG requires[3]

$$T_{\mathsf{grad}} = O\Big(\big(n+\frac{n^{3/4}\sqrt{L}}{\sqrt{\sigma}}\big) \log^2 \frac{1}{\varepsilon}\Big) \text{ if } \sigma > 0 \quad \text{or} \quad T_{\mathsf{grad}} = O\Big(\big(n+\frac{n^{3/4}\sqrt{L}}{\sqrt{\varepsilon}}\big) \log \frac{1}{\varepsilon}\Big) \text{ if } \sigma = 0 \quad . \quad (1.3)$$

Unfortunately, the practicality of AccSVRG remains somewhat unsettled. To mention a few issues:

- Since error propagates, one needs to run each SVRG until a very accurate point is obtained.
- To optimize the complexity, one needs to terminate each call of SVRG at a different accuracy.
- One needs to tune three parameters: (1) the regularizer weight $L'$, (2) the learning rate of SVRG, and (3) the weight of the momentum.

When all of these factors are putting together, the practical performance of AccSVRG may be even worse than the non-accelerated SVRG.[4]

---

[2]Both reductions are based on an outer-inner loop structure first proposed by Shalev-Shwartz and Zhang [28]. The application of Catalyst to solving Problem (1.1) first appeared in [14, 27] in the context of PCA.

[3]For why these logarithmic factors show up, we refer readers to the journal version of Catalyst [19]. In particular, one of the two log factors in the case of $\sigma > 0$ is because SVRG is a randomized algorithm.

[4]This is so already in the easier case where each $f_i(x)$ is convex [2]. In this simpler case, direct acceleration (without applying the Catalyst reduction) is more practical and already known (see `Katyusha` [2] and `DASVRDA` [21]).

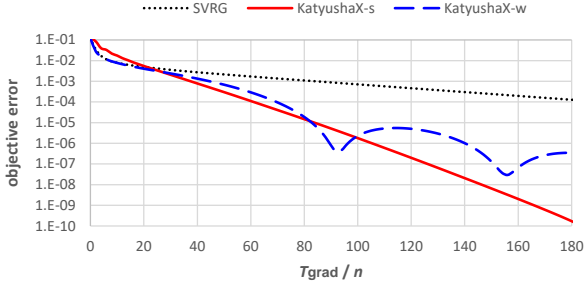
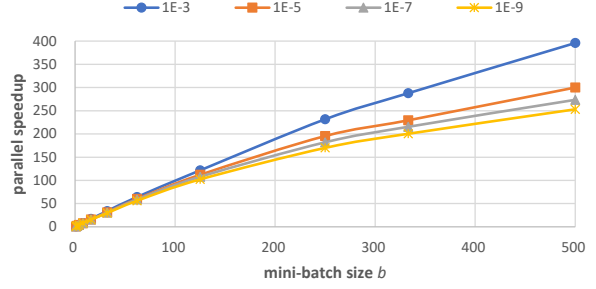

(a) `SVRG` vs. `KatyushaX`$^s$ vs. `KatyushaX`$^w$    (b) mini-batch performance of `KatyushaX`$^s$

Figure 1: A simple illustration on minimizing $f(x) = \frac{1}{2}(\mu I - BB^\top)$ where $B \in \mathbb{R}^{1000 \times 1000}$ is a random $\pm 1$ matrix, and $\mu = \lambda_1(BB^\top) + 0.5\big(\lambda_1(BB^\top) - \lambda_2(BB^\top)\big)$. Such $f(x)$ is a typical instance in stochastic PCA [14].
**Remark 1.** In `SVRG`, the best learning rate is $\eta = 0.4/L$ after tuning.
**Remark 2.** We used $\eta = 0.4/L$ for `KatyushaX`$^w$. We used $\eta = 0.4/L$ and $\tau = 0.1$ for `KatyushaX`$^s$.
**Remark 3.** In the mini-batch experiment, we used $\eta = \frac{0.4b}{L}$ and $\tau = 0.1$. The parallel speed-up is in terms of achieving objective error $10^{-3}, 10^{-5}, 10^{-7}, 10^{-9}$.

## 1.3 Our Main Result

In this paper, we propose a new method `KatyushaX` which, copying the original SVRG method but adding only one additional line, achieves the accelerated stochastic convergence rate. We give two different specifications, `KatyushaX`$^s$ and `KatyushaX`$^w$, where

- `KatyushaX`$^s$ needs a momentum parameter in addition to the learning rate for SVRG; and
- `KatyushaX`$^w$ needs no additional parameter whatsoever on top of SVRG.

We explain how they work below, and Figure 1(a) gives a quick performance comparison between `SVRG`, `KatyushaX`$^s$ and `KatyushaX`$^w$ on some synthetic dataset.

We first recall how SVRG works. Each *epoch* of SVRG consists of $n$ iterations. Each epoch starts with a point $w_0$ (known as the *snapshot*) where the full gradient $\nabla f(w_0)$ is computed exactly. Then, in each iteration $t = 0, 1, \ldots, n-1$ of this epoch, SVRG updates $w_{t+1} \leftarrow w_t - \eta \widetilde{\nabla}_t$ where the gradient estimator $\widetilde{\nabla}_t = \nabla f_i(w_t) - \nabla f_i(w_0) + \nabla f(w_0)$ for some random $i \in [n]$, and $\eta > 0$ is the learning rate.

If we summarize the above one epoch process of SVRG as $\mathtt{SVRG}^{\mathtt{1ep}}(w_0, \eta) \stackrel{\mathrm{def}}{=} w_n$, then

- The classical SVRG method can be described by the iterative update

$$\boxed{x_{k+1} = \mathtt{SVRG}^{\mathtt{1ep}}(x_k, \eta) \ .}$$

- Our `KatyushaX`$^s$, parameterized by a momentum parameter $\tau \in (0, 1)$, is

$$\boxed{y_k \leftarrow \mathtt{SVRG}^{\mathtt{1ep}}(x_k, \eta) \quad \text{and} \quad x_{k+1} \leftarrow \frac{\frac{3}{2}y_k + \frac{1}{2}x_k - (1-\tau)y_{k-1}}{1+\tau}} \ .$$

- Our `KatyushaX`$^w$ is

$$\boxed{y_k \leftarrow \mathtt{SVRG}^{\mathtt{1ep}}(x_k, \eta) \quad \text{and} \quad x_{k+1} \leftarrow \frac{(3k+1)y_k + (k+1)x_k - (2k-2)y_{k-1}}{2k+4}} \ .$$

*Remark 1.3.* When choosing $\tau = 1/2$, `KatyushaX`$^s$ is exactly identical to SVRG and $y_k \equiv x_{k+1}$.

*Remark 1.4.* In `KatyushaX`$^s$, if we replace $\frac{3}{2}y_k + \frac{1}{2}x_k$ with $2y_k$, then the update becomes $x_{k+1} \leftarrow$

$y_k + \frac{1-\tau}{1+\tau}(y_k - y_{k-1})$. This corresponds to a classical momentum scheme by Nesterov [22]. The smaller $\tau > 0$ is the "stronger" the momentum behaves.

*Remark* 1.5. `KatyushaX`$^{\text{w}}$ is in fact `KatyushaX`$^{\text{s}}$ with $\tau = \frac{2}{k+2}$ decreasing in $k$. See Fact 4.5.

Our main theorems are the following:

**Theorem 1** (informal). *If $f(x)$ is $\sigma$-strongly convex and each $f_i(x)$ is $L$-smooth, then `KatyushaX`$^{\text{s}}$ with $\eta = \Theta\left(\frac{1}{\sqrt{nL}}\right)$ and $\tau = \min\{\frac{1}{2}, \Theta(\frac{n^{1/4}\sqrt{\sigma}}{\sqrt{L}})\}$ outputs a point $x$ with $\mathbb{E}[f(x) - f(x^*)] \leq \varepsilon$ using*

$$T_{\text{grad}} = O\Big(\big(n + \frac{n^{3/4}\sqrt{L}}{\sqrt{\sigma}}\big) \log \frac{1}{\varepsilon}\Big) \text{ stochastic gradient computations.}$$

**Theorem 2** (informal). *If $f(x)$ is convex and each $f_i(x)$ is $L$-smooth, then `KatyushaX`$^{\text{w}}$ with $\eta = \Theta\left(\frac{1}{\sqrt{nL}}\right)$ outputs a point $x$ with $\mathbb{E}[f(x) - f(x^*)] \leq \varepsilon$ using*

$$T_{\text{grad}} = O\Big(n + \frac{n^{3/4}\sqrt{L}}{\sqrt{\varepsilon}}\Big) \text{ stochastic gradient computations.}$$

In sum, we have not only tightened the complexity bounds by removing a logarithmic factor each (comparing to AccSVRG (1.3)), but also obtained a much *simpler, practical* acceleration (i.e., momentum) scheme for minimizing sum-of-nonconvex functions stochastically.

### 1.4 Our Side Results

To demonstrate the strength of `KatyushaX`, we prove our main theorem in several more general settings.

**(1) Upper and Lower Smoothness.** For non-convex functions, its upper and lower smoothness parameters (i.e., maximum eigenvalue vs. negated minimum eigenvalue of Hessian) may be very different. This is especially true for all PCA and SVD related applications [7, 8, 14], where $f_i(x) = \frac{\mu}{2}\|x\|^2 - \langle a_i, x\rangle^2$ so its upper smoothness is $\mu$ and lower smoothness is $2\|a_i\|^2$. (It usually happens that $\|a_i\|^2 \gg \mu$.)

Assuming $f_i(x)$ is $\ell_1$-upper and $\ell_2$-lower smooth (and $\ell_2 \geq \ell_1$), it is known that by simply changing the learning rate of SVRG, its runs in a worst-case complexity proportional to $\sqrt{\ell_1 \ell_2}$ instead of $L$ [11].

We show `KatyushaX` enjoys this speed up as well.

`KatyushaX` runs in a complexity proportional to $(\ell_1 \ell_2)^{1/4}$ instead of $L^{1/2}$.

**(2) Composite Minimization.** Consider objective $F(x) = \psi(x) + \frac{1}{n}\sum_{i=1}^{n} f_i(x)$ where $\psi(x)$ is some proper convex (not necessarily smooth) function, usually referred to as the proximal term.[5] Then, most stochastic gradient methods can be extended to minimize composite objectives, if we replace the update $w_{t+1} \leftarrow w_t - \eta \widetilde{\nabla}_t$ with $w_{t+1} \leftarrow \arg\min_{z \in \mathbb{R}^d} \{\frac{1}{2\eta}\|z - w_t\|_2^2 + \langle \widetilde{\nabla}_t, z\rangle + \psi(z)\}$.

`KatyushaX` also extends to the composite minimization setting.

**(3) Parallelism / Mini-batch.** Instead of using a single stochastic gradient $\nabla f_i(\cdot)$ per iteration, for any stochastic method, one can replace it with the average of $b$ stochastic gradients $\frac{1}{b}\sum_{i \in S} \nabla f_i(\cdot)$

---
[5]Examples of proximal terms include $\psi(x) = \|x\|_1$ or $\psi(x) = \begin{cases} 0, & x \in \mathcal{X}; \\ +\infty, & x \notin \mathcal{X}. \end{cases}$ for some convex set $\mathcal{X} \subseteq \mathbb{R}^d$.

where $S$ is a random subset of $[n]$ with cardinality $b$. This is known as the *mini-batch* technique and it allows the stochastic gradients to be computed in a distributed manner, using up to $b$ processors.

Our KatyushaX methods extends to this mini-batch setting too. Using mini-batch size $b$,

$$\text{the worst-case \# parallel iterations of } \texttt{KatyushaX} \text{ reduces by } \begin{cases} O(b), & \text{if } b \leq \sqrt{n}; \\ O(\sqrt{b}), & \text{if } b \in [\sqrt{n}, n]. \end{cases}$$

Therefore, at least for small $b \in \{1, 2, \ldots, \lceil \sqrt{n} \rceil\}$, KatyushaX enjoys a *linear speed-up* in the parallel worst-case running time. We do not find such result recorded before this work.

**(4) Non-Uniform Sampling.** When functions $f_i(x)$ are of non-uniform hardness (say, with different smoothness parameters), instead of sampling each function $f_i(x)$ uniformly at random, one can sample $i$ with a probability proportional to its "hardness." This can improve the performance of stochastic gradient methods.

<p style="text-align:center">KatyushaX also enjoys non-uniform sampling benefits.</p>

If each function $f_i(x)$ is $L_i$-smooth, one can sample $i$ with probability proportional to $L_i^2$. If we denote by $\overline{L} = (\sum_i L_i^2/n)^{1/2}$, then the worst-case complexities can be improved to depend on $\overline{L}$ instead of $\max_i\{L_i\}$.

## 1.5 Other Related Works

Since sum-of-nonconvex optimization is closely related to PCA, let us mention the most standard variants of PCA and their relationships to Problem (1.1).

**Offline Stochastic PCA.** In the offline setting, we assume $A = \frac{1}{n}\sum_{i=1}^n a_i a_i^\top$ and a stochastic method can compute $\langle a_i, x \rangle$ for some vector $x$ in each iteration.

To approximate the top eigenvector of $A$ (i.e., 1-PCA), the first variance reduction method is by Shamir [29] and does not need sum-of-nonconvex optimization. Unfortunately, his method is not gap-free[6] and not accelerated. Garber et al. [14] obtained a stochastic, gap-free, and accelerated method by reducing 1-PCA to minimizing *sum-of-nonconvex functions* using shift-and-invert. To this date, this seems to be the only approach to obtain a stochastic *and* accelerated method for 1-PCA.

To approximate the top $k$ eigenvectors of $A$ (i.e., $k$-PCA), the first variance reduction method is by Shamir [30] and does not need sum-of-nonconvex optimization. His method is not gap-free, not accelerated, and has a slow worst-case complexity. Allen-Zhu and Li [7] obtained a stochastic, gap-free, and accelerated method for $k$-PCA by reducing the problem to minimizing *sum-of-nonconvex functions*. To this date, this seems to be the only approach to obtain a stochastic *and* accelerated method for $k$-PCA.

**Online Stochastic PCA.** In the online setting, we assume $A = \mathbb{E}_i[a_i a_i^\top]$ where there may be infinitely many vectors $a_i$ so the complexity of the stochastic method cannot depend on $n$.

In the case of online 1-PCA, the optimal algorithm is Oja's method [24], whose first optimal analysis was due to Jain et al. [15] and first optimal *gap-free* analysis was due to Allen-Zhu and Li [9]. In the case of online $k$-PCA, the optimal algorithm is a block variant of Oja's method, whose first optimal analysis was due to Allen-Zhu and Li [8].

In the online stochastic setting, due to information-theoretic lower bounds [8, 9], one cannot apply variance reduction or acceleration to improve the worst-case complexity.

**Online Adversarial PCA.** In an online learning scenario where the player chooses a unit vector $v_t$ at round $t$ and the adversary chooses a matrix $A_t$, one can design regret-minimizing strategy for the player in terms of maximizing $\sum_t v_t^\top A_t v^t$. In this game, the regret-optimal strategy for the player is follow-the-regularized-leader (FTRL) but it runs very slow; an efficient strategy for the player is follow-the-perturbed-leader (FTPL) but it gives a poor regret [20]. The recent result follow-the-compressed-leader (FTCL) gives a strategy that is both regret near-optimal and efficiently computable [9]. Both strategies FTCL and FTPL rely on minimizing *sum-of-nonconvex functions*.

---

[6]We say a method is gap-free if it does not need an eigengap assumption between the top two eigenvalues.

**Stochastic Nonconvex Optimization.** In the harder problem where $f(x) = \frac{1}{n}\sum_{i=1}^n f_i(x)$ is also non-convex, variance reduction is also proven useful, both in terms of finding approximate stationary points and approximate local minima.

In the finite-sum case (i.e., when $n$ is finite), the SVRG method finds an $\varepsilon$-approximate stationary point in $T_{\mathsf{grad}} = O(n^{2/3}/\varepsilon^2)$ where in contrast SGD needs $O(1/\varepsilon^4)$ and full gradient descent needs $O(n/\varepsilon^2)$.[7] This was first shown independently by Reddi et al. [25] and Allen-Zhu and Hazan [5]. To find an $\varepsilon$-approximate local minima, one needs an additional second-order smoothness assumption, and the two independent works Agarwal et al. [1], Carmon et al. [12] need $T_{\mathsf{grad}} = O\big(\frac{n}{\varepsilon^{1.5}} + \frac{n^{3/4}}{\varepsilon^{1.75}}\big)$. Both these two algorithms rely on minimizing *sum-of-nonconvex functions*.

In the online case (i.e., when $n$ is infinite), the SCSG method (which is a variant of SVRG) finds an $\varepsilon$-approximate stationary point in $T_{\mathsf{grad}} = O(1/\varepsilon^{3.333})$, and this was first discovered by Lei et al. [17]. To find an $\varepsilon$-approximate local minima, one needs an additional second-order smoothness assumption and the convergence rate can be improved to $T_{\mathsf{grad}} = O(1/\varepsilon^{3.25})$ [3]. Neither algorithm relies on sum-of-nonconvex functions, and momentum is not known to be helpful in the online setting.

**Relationship to Katyusha.** We have borrowed the algorithm name from [2], where the author obtained a *direct* accelerated stochastic method Katyusha for minimizing a function $f(x)$ that is a finite average of *convex* functions $f_i(x)$. The two algorithms are different:

- Katyusha applies a momentum step every iteration, but KatyushaX applies a momentum step every epoch (i.e., every $n$ iterations).
- KatyushaX applies to a more general class of sum-of-nonconvex functions than Katyusha.
- Katyusha gives a better complexity than KatyushaX when restricted to convex $f_i(x)$.

The two works also share some similarity.

- Both works provably and directly add momentum to a stochastic method. This can have practical impacts.
- Both works introduces some "negative momentum" on top of SVRG. In every iteration of Katyusha, the point retracts towards the most recent snapshot (this is achieved by a three-point linear coupling [2]). In KatyushaX, after each epoch we applied $x_{k+1} \leftarrow \frac{\frac{3}{2}y_k + \frac{1}{2}x_k - (1-\tau)y_{k-1}}{1+\tau}$ which is different from the classical update $x_{k+1} \leftarrow \frac{2y_k - (1-\tau)y_{k-1}}{1+\tau}$. This can also be viewed as retracting $y_k$ towards the most recent snapshot $x_k$.

## 1.6 Roadmap

We give preliminaries and formalize the problem in Section 2. We give a new analyze of one epoch of SVRG in Section 3, and then accelerate it in Section 4. We prove our main theorems in the uniform sampling setting, and defer the non-uniform variant to Section 5.

---
[7]In this high-level summary, we have hidden the smoothness and variance parameters inside the big-$O$ notion.

## 2 Preliminaries

Throughout this paper, we denote by $\|\cdot\|$ the Euclidean norm. We use $i \in_R [n]$ to denote that $i$ is generated from $[n] = \{1, 2, \ldots, n\}$ uniformly at random. We denote by $\nabla f(x)$ the gradient of function $f$ if it is differentiable, and $\partial f(x)$ any subgradient if $f$ is only Lipschitz continuous. Recall some definitions on strong convexity and smoothness — they have other equivalent definitions, see textbook [22].

**Definition 2.1.** *For a function $f \colon \mathbb{R}^d \to \mathbb{R}$,*
- *$f$ is $\sigma$-strongly convex if $\forall x, y \in \mathbb{R}^d$, it satisfies $f(y) \geq f(x) + \langle \partial f(x), y - x \rangle + \frac{\sigma}{2}\|x-y\|^2$.*
- *$f$ is $L$-Lipschitz smooth (or $L$-smooth for short) if $\forall x, y \in \mathbb{R}^d$, $\|\nabla f(x) - \nabla f(y)\| \leq L\|x - y\|$.*
- *$f$ is $(\ell_1, \ell_2)$-smooth if for every $\forall x, y \in \mathbb{R}^d$*

$$-\tfrac{\ell_2}{2}\|x-y\|^2 \leq f(y) - \big(f(x) + \langle \nabla f(x), y - x \rangle\big) \leq \tfrac{\ell_1}{2}\|x-y\|^2 \ .$$

*A convex $L$-smooth function is $(L, 0)$-smooth; a non-convex $L$-smooth function is $(L, L)$-smooth.*

### 2.1 Problem Formulation

Throughout this paper, we minimize the following convex stochastic composite objective over $x \in \mathbb{R}^d$:

$$F(x) = \psi(x) + f(x) \stackrel{\text{def}}{=} \psi(x) + \frac{1}{n}\sum_{i \in [n]} f_i(x) \ , \tag{2.1}$$

where

1. $\psi(x)$ is proper convex (a.k.a. the proximal term).
2. Each $f_i(x)$ is differentiable and $(\ell_1, \ell_2)$-smooth (for $\ell_2 \geq \ell_1$); and $f(x)$ is convex and $L$-smooth.
3. $f(x)$ is $\sigma_f$-strongly convex and $\psi(x)$ is $\sigma_\psi$-strongly convex,

*Remark* 2.2. Both $\sigma_f, \sigma_\psi$ can be zero, and only $\sigma \stackrel{\text{def}}{=} \sigma_f + \sigma_\psi$ needs to be known by the algorithm. Also, we use the $(\ell_1, \ell_2)$-smoothness notion to provide the tightest bound possible. In a first reading, one can simply view "$\ell_1 = \ell_2 = L$" and this case is sufficiently interesting.

In the rest of the paper, we define $T_{\mathsf{grad}}$, the gradient complexity, as the number of computations of $\nabla f_i(x)$; and in the mini-batch setting, we define $T_{\mathsf{iter}}$ as the number of parallel iterations (where in each iteration we can compute $b$ stochastic gradients in parallel).

### 2.2 Geometric Distribution

We say $N$ follows from geometric distribution with parameter $p \in (0, 1]$, denoted by $\mathsf{Geom}(p)$, if $N = k$ with probability $(1-p)^k p$ for each $k \in \{0, 1, 2, \ldots\}$. We know $\mathbb{E}[N] = \frac{1-p}{p}$. We have:

**Fact 2.3.** *Given any sequence $D_0, D_1, \ldots$ of reals, if $N \sim \mathsf{Geom}(p)$, then*
- $\mathbb{E}_N[D_N - D_{N+1}] = \frac{p}{1-p}\mathbb{E}[D_0 - D_N]$, *and*
- $\mathbb{E}_N[D_N] = (1-p)\mathbb{E}[D_{N+1}] + pD_0$

### 2.3 Proximal Mirror Descent

We recall the following classical lemma for proximal mirror descent, and its proof is classical (see Appendix A).

**Algorithm 1** $\texttt{SVRG}^{\texttt{1ep}}(F, w_0, b, \eta)$

**Input:** $F(x) = \psi(x) + \frac{1}{n}\sum_{i=1}^{n} f_i(x)$, starting vector $w_0$, mini-batch size $b \in [n]$, learning rate $\eta > 0$.
**Output:** vector $w^+$.
 1: $m \leftarrow \min\{\lceil \frac{n}{b} \rceil, 2\}$; $M \sim \mathsf{Geom}(\frac{1}{m})$;  ⋄ $\mathbb{E}[M] = m - 1$, *in practice just let* $M \leftarrow m - 1$
 2: $\mu = \nabla f(w_0)$;
 3: **for** $t \leftarrow 0$ **to** $M$ **do**
 4:     Let $S_t$ be $b$ i.i.d. uniform random indices from $[n]$ with replacement;
 5:     $\widetilde{\nabla}_t \leftarrow \mu + \frac{1}{b}\sum_{i \in S_t}\left(\nabla f_i(w_t) - \nabla f_i(w_0)\right)$;
 6:     $w_{t+1} = \arg\min_{y \in \mathbb{R}^d}\left\{\psi(y) + \frac{1}{2\eta}\|y - w_t\|^2 + \langle\widetilde{\nabla}_t, y\rangle\right\}$
 7: **end for**
 8: **return** $w^+ \leftarrow w_{M+1}$.

**Lemma 2.4.** *If $h(\cdot)$ is proper convex and $\sigma$-strongly convex and $z_{k+1} = \arg\min_{z \in \mathbb{R}^d}\{\frac{1}{2\alpha}\|z - z_k\|^2 + \langle\xi, z\rangle + h(z)\}$, then for every $u \in \mathbb{R}^d$, we have*

$$\langle\xi, z_k - u\rangle + h(z_{k+1}) - h(u) \leq \langle\xi, z_k - z_{k+1}\rangle + \frac{\|u - z_k\|^2}{2\alpha} - \frac{(1+\sigma\alpha)\|u - z_{k+1}\|^2}{2\alpha} - \frac{\|z_k - z_{k+1}\|^2}{2\alpha}$$

$$\leq \frac{\alpha}{2}\|\xi\|^2 + \frac{\|u - z_k\|^2}{2\alpha} - \frac{(1+\sigma\alpha)\|u - z_{k+1}\|^2}{2\alpha} \ .$$

## 3 New Analysis of SVRG

In this section, we propose a *new* analysis for one epoch of SVRG specialized sum-of-nonconvex optimization (2.1).

Recall the original design of SVRG is due to [16, 32] and its the first proximal analysis is due to [31]. The first sum-of-nonconvex analysis for SVRG is due to [27] and its proximal analysis seems to be due to [11]. Our analysis is going to be different from all of these cited results.

We formally specify the one epoch instance of SVRG in Algorithm 1 and call it $\texttt{SVRG}^{\texttt{1ep}}$. To provide the most general statement, we have allowed a mini-batch size $b \in [n]$ to be specified as input to $\texttt{SVRG}^{\texttt{1ep}}$ (and thus an epoch corresponds to $\approx \frac{n}{b}$ iterations), and have allowed a proximal term $\psi(x)$ to exist in the stochastic gradient updates.

We define $\texttt{SVRG}^{\texttt{1ep}}$ so that it randomly stops in $M$ iterations where $M$ follows from a geometric distribution whose expectation equals $\approx \frac{n}{b}$. This is for the purpose of simplifying analysis (motivated by Lei et al. [17]). Random stopping is not necessary in practice.[8]

Our main theorem of this section is:

**Theorem 3.1.** *Let $b \in [n]$ be the mini-batch size and $m = \min\{\lceil\frac{n}{b}\rceil, 2\}$ be the epoch length of $\texttt{SVRG}^{\texttt{1ep}}$. If $w^+ = \texttt{SVRG}^{\texttt{1ep}}(F, w_0, b, \eta)$ where $\eta \leq \min\left\{\frac{1}{2L}, \frac{1}{2\sqrt{\ell_1\ell_2 m/b}}\right\}$, then*

$$\forall u \in \mathbb{R}^d: \quad \mathbb{E}\big[F(w^+) - F(u)\big] \leq \mathbb{E}\bigg[-\frac{1}{4m\eta}\|w^+ - w_0\|^2 + \frac{\langle w_0 - w^+, w_0 - u\rangle}{m\eta} - \frac{\sigma}{4}\|w^+ - u\|^2\bigg] \ .$$

**Interpretation.** We can interpret Theorem 3.1 as gradient descent for the following reason. Denoting by $\mathcal{G} = \frac{w_0 - w^+}{m\eta}$, then it satisfies

$$\mathbb{E}\big[F(w^+) - F(u)\big] \leq \mathbb{E}\bigg[-\frac{m\eta}{4}\|\mathcal{G}\|^2 + \langle\mathcal{G}, w_0 - u\rangle - \frac{\sigma}{4}\|w^+ - u\|^2\bigg] \ . \tag{3.1}$$

---
[8]In practice, one can simply choose $M = \frac{n}{b}$ as fixed. This is how our experiments in Figure 1 are performed.

In comparison, if a full proximal gradient descent with step length $\frac{1}{L}$ is applied $y^+ = \arg\min_z \{\psi(z) + \frac{L}{2}\|z-y\|^2 + \langle \nabla f(y), z\rangle\}$, and denote by $\mathcal{G} = L(y - y^+)$ the so-called *gradient mapping*, then classical theory essentially tells us:[9]

$$\mathbb{E}\big[F(y^+) - F(u)\big] \leq \mathbb{E}\Big[ -\frac{1}{2L}\|\mathcal{G}\|^2 + \langle \mathcal{G}, y-u\rangle - \frac{\sigma}{2}\|y^+ - u\|^2 \Big] \ . \tag{3.2}$$

By comparing (3.1) and (3.2), we see that up to a constant factor of 2, SVRG[1ep] can be viewed as a full gradient descent with a "virtual" step length $m\eta$.

We shall later use SVRG[1ep] (and thus Theorem 3.1) in a black-box way to obtain both accelerated methods in Section 4 and non-accelerated ones in Appendix D.

### 3.1 Proof Details

Using the convexity and smoothness of our objective, as well as the definition of our stochastic gradient step, we obtain the following standard lemma (see Appendix ??):

**Lemma 3.2.** *If $w_{t+1} = \arg\min_{y\in\mathbb{R}^d}\{\frac{1}{2\eta}\|y - w_t\|^2 + \psi(y) + \langle \widetilde{\nabla}_t, y\rangle\}$ for some random vector $\widetilde{\nabla}_t \in \mathbb{R}^d$ satisfying $\mathbb{E}[\widetilde{\nabla}_t] = \nabla f(w_t)$, then for every $u \in \mathbb{R}^d$, we have*

$$\mathbb{E}\big[F(w_{t+1}) - F(u)\big] \leq \mathbb{E}\Big[\frac{\eta}{2(1-\eta L)}\|\widetilde{\nabla}_t - \nabla f(w_t)\|^2 + \frac{(1-\sigma_f\eta)\|u - w_t\|^2 - (1+\sigma_\psi\eta)\|u - w_{t+1}\|^2}{2\eta}\Big] \ .$$

The next lemma is one of our main contributions, and can be proved by telescoping Lemma 3.2 (see Appendix B).

**Lemma 3.3.** *If the gradient estimator $\widetilde{\nabla}_t$ satisfies $\mathbb{E}[\widetilde{\nabla}_t] = \nabla f(w_t)$ and $\mathbb{E}\big[\|\widetilde{\nabla}_t - \nabla f(w_t)\|^2\big] \leq Q\|w_0 - w_t\|^2$ for every $t$ for some universal constant $Q > 0$, then as long as $\eta \leq \min\{\frac{1}{2L}, \frac{1}{2\sqrt{Qm}}\}$, it satisfies*

$$\forall u \in \mathbb{R}^d: \quad \mathbb{E}\big[F(w^+) - F(u)\big] \leq \mathbb{E}\Big[ -\frac{1}{4m\eta}\|w^+ - w_0\|^2 + \frac{\langle w_0 - w^+, w_0 - u\rangle}{m\eta} - \frac{\sigma_f + \sigma_\psi}{4}\|w^+ - u\|^2 \Big] \ .$$

Our next lemma (to be proved in Appendix C) shows that one can choose $Q = \frac{\ell_1\ell_2}{b}$ in Lemma 3.3, and this concludes the proof of Theorem 3.1.

**Lemma 3.4.** *If each $f_i(x)$ is $(\ell_1, \ell_2)$-smooth for $\ell_2 \geq \ell_1$, and the gradient estimator*

$$\widetilde{\nabla}_t = \mu + \tfrac{1}{b}\sum_{i\in S_t}\big(\nabla f_i(w_t) - \nabla f_i(w_0)\big)$$

*where $S_t$ consists of $b$ i.i.d. uniform random indices from $[n]$ with replacement. Then, $\mathbb{E}_{S_t}[\widetilde{\nabla}_t] = \nabla f(w_t)$ and $\mathbb{E}_{S_t}\big[\|\widetilde{\nabla}_t - \nabla f(w_t)\|^2\big] \leq \frac{\ell_1\ell_2}{b}\|w_t - w_0\|^2$ .*

## 4 Acceleration

We give some intuitions why direct acceleration is possible using SVRG[1ep] in the context of $\sigma > 0$ — the $\sigma = 0$ case is similar.

Recall Nesterov's momentum turns gradient descent with step length $\frac{1}{L}$ into an algorithm with $O\big((1 + \sqrt{L/\sigma})\log\frac{1}{\varepsilon}\big)$ iterations. Since SVRG[1ep] can be viewed as gradient descent with a "virtual" step length $m\eta$ (see Section 3), one should hope that perhaps Nesterov's momentum can also turn

---

[9]See for instance [31, Lemma 3.7]. In fact, the exact statement from [31, Lemma 3.7] is by replacing $\frac{\sigma}{2}\|y^+ - u\|^2$ with $\frac{\sigma_\psi}{2}\|y^+ - u\|^2 + \frac{\sigma_f}{2}\|y - u\|^2$. We ignore this subtle issue in this high-level interpretation.

`SVRG^{1ep}` into an algorithm with $O\big((1+\sqrt{1/\eta m\sigma})\log\frac{1}{\varepsilon}\big)$ iterations. Since each call to `SVRG^{1ep}` requires computing $O(n)$ stochastic gradients, this "wishful thinking" gives

$$T_{\mathsf{grad}} = n \times O\Big(\big(1 + \sqrt{1/\eta m\sigma}\big)\log\frac{1}{\varepsilon}\Big) = O\Big(\big(n + \frac{\sqrt{Lbn} + (\ell_1\ell_2)^{1/4}n^{3/4}}{\sqrt{\sigma}}\big)\log\frac{1}{\varepsilon}\Big) \ .$$

This is exactly what we hope to obtain. In the case of $b=1$ and $\ell_1 = \ell_2 = L$, it gives the desired complexity $T_{\mathsf{grad}} = O\big(\big(n + \frac{\sqrt{L}n^{3/4}}{\sqrt{\sigma}}\big)\log\varepsilon^{-1}\big)$ in Theorem 1.

Unfortunately, some technical issue arises. To the best of our knowledge, the constant 4 in (3.1) —as opposed to 2 in the true gradient descent (3.2)— actually prevents the classical Nesterov's momentum from being applied in the worst case. In symbols, this means the standard choice $x_{k+1} = \frac{2y_k - (1-\tau)y_{k-1}}{1+\tau}$ does not seem to work.

In `KatyushaX^s`, we use $x_{k+1} = \frac{\frac{3}{2}y_k + \frac{1}{2}x_k - (1-\tau)y_{k-1}}{1+\tau}$. Intuitively, this replaces one copy of $y_k$ with $\frac{y_k+x_k}{2}$ in the original Nesterov's momentum. Since $x_k$ is the last point where the full gradient is exactly computed, retracting towards $x_k$ stabilizes the algorithm.

On the mathematical side, this new choice of momentum is motivated by the linear-coupling analysis of accelerated methods [10]. In their language, we consider the following class of accelerated methods.

---
**General Framework**

Starting from $z_0 = y_0 = x_0$, then in each iteration $k = 0, 1, \ldots, K-1$, (4.1)

- $x_{k+1} \leftarrow \tau_k z_k + (1-\tau_k)y_k$ for some $\tau_k \in [0,1]$;
- $y_{k+1} = \mathtt{SVRG}^{\mathtt{1ep}}(F, x_{k+1}, b, \eta)$ and let $\mathcal{G}_{k+1} = \frac{x_{k+1} - y_{k+1}}{m\eta}$ be the gradient mapping;
- $z_{k+1} \leftarrow \arg\min_{z\in\mathbb{R}^d}\big\{\frac{1}{2\alpha_{k+1}}\|z-z_k\|^2 + \langle \mathcal{G}_{k+1}, z\rangle + \frac{\sigma}{4}\|z-y_{k+1}\|^2\big\}$ for some $\alpha_{k+1} > 0$.
---

We will show that `KatyushaX^s` and `KatyushaX^w` are two instantiations of this general class of methods, after choosing $\alpha_{k+1}$ and $\tau_k$ appropriately.

## 4.1 The Core Lemma

Our next lemma is the core lemma to show accelerated convergence rates. It is based on the linear-coupling idea of [10, Lemma 4.3], which combines analyses of gradient descent and mirror descent. In particular, we view `SVRG^{1ep}` as gradient descent (see Section 3), and view the update sequence on $\{z_k\}_k$ as mirror descent (see Lemma 2.4). We include its half-paged proof in Appendix ??.

**Lemma 4.1.** *If $\alpha_{k+1} \leq \frac{m\eta}{2\tau_k}$ in the General Framework (4.1), then for every $u \in \mathbb{R}^d$:*

$$\frac{\alpha_{k+1}}{\tau_k}\big(F(y_{k+1}) - F(u)\big) \leq \frac{(1-\tau_k)\alpha_{k+1}}{\tau_k}\big(F(y_k) - F(u)\big) + \Big(\frac{1}{2}\|z_k - u\|^2 - \frac{1+\alpha_{k+1}\sigma/2}{2}\|z_{k+1} - u\|^2\Big)$$

*Proof.* Denoting by $\mathcal{G}_{k+1} = \frac{x_{k+1}-y_{k+1}}{m\eta}$, then Lemma 3.3 implies

$$\forall u \in \mathbb{R}^d: \quad \mathbb{E}\big[F(y_{k+1}) - F(u)\big] \leq \mathbb{E}\Big[-\frac{m\eta}{4}\|\mathcal{G}_{k+1}\|^2 + \langle \mathcal{G}_{k+1}, x_{k+1} - u\rangle - \frac{\sigma}{4}\|y_{k+1} - u\|^2\Big] \ . \quad (4.2)$$

Therefore,

$$\alpha_{k+1}\big(F(y_{k+1}) - F(u)\big) \stackrel{\text{①}}{\leq} \alpha_{k+1}\langle \mathcal{G}_{k+1}, x_{k+1} - u\rangle - \frac{\alpha_{k+1} m\eta}{4}\|\mathcal{G}_{k+1}\|^2 - \frac{\alpha_{k+1}\sigma}{4}\|y_{k+1} - u\|^2$$

$$= \alpha_{k+1}\langle \mathcal{G}_{k+1}, x_{k+1} - z_k\rangle - \frac{\alpha_{k+1} m\eta}{4}\|\mathcal{G}_{k+1}\|^2 + \alpha_{k+1}\Big(\langle \mathcal{G}_{k+1}, z_k - u\rangle - \frac{\sigma}{4}\|y_{k+1} - u\|^2\Big)$$

$$\stackrel{\text{②}}{=} \frac{(1-\tau_k)\alpha_{k+1}}{\tau_k}\langle \mathcal{G}_{k+1}, y_k - x_{k+1}\rangle - \frac{\alpha_{k+1} m\eta}{4}\|\mathcal{G}_{k+1}\|^2 + \alpha_{k+1}\Big(\langle \mathcal{G}_{k+1}, z_k - u\rangle - \frac{\sigma}{4}\|y_{k+1} - u\|^2\Big)$$

$$\stackrel{\text{③}}{\leq} \frac{(1-\tau_k)\alpha_{k+1}}{\tau_k}\Big(F(y_k) - F(y_{k+1}) - \frac{m\eta}{4}\|\mathcal{G}_{k+1}\|^2\Big) - \frac{\alpha_{k+1} m\eta}{4}\|\mathcal{G}_{k+1}\|^2$$

$$+ \Big(\frac{\alpha_{k+1}^2}{2}\|\mathcal{G}_{k+1}\|^2 + \frac{1}{2}\|z_k - u\|^2 - \frac{1+\alpha_{k+1}\sigma/2}{2}\|z_{k+1} - u\|^2\Big)$$

Above, inequality ① applies (4.2); equality ② uses the definition $x_{k+1} \leftarrow \tau_k z_k + (1-\tau_k)y_k$; and inequality ③ uses $z_{k+1} = \arg\min_{z\in\mathbb{R}^d}\big\{\frac{1}{2\alpha_{k+1}}\|z - z_k\|^2 + \langle \frac{x_{k+1} - y_{k+1}}{m\eta}, z\rangle + \frac{\sigma}{4}\|z - y_{k+1}\|^2\big\}$ and Lemma 2.4. Finally, rearranging terms, we have

$$\frac{\alpha_{k+1}}{\tau_k}\big(F(y_{k+1}) - F(u)\big) \leq \frac{(1-\tau_k)\alpha_{k+1}}{\tau_k}\big(F(y_k) - F(u)\big) + \Big(\frac{\alpha_{k+1}^2}{2} - \frac{\alpha_{k+1} m\eta}{4\tau_k}\Big)\|\mathcal{G}_{k+1}\|^2$$

$$+ \Big(\frac{1}{2}\|z_k - u\|^2 - \frac{1+\alpha_{k+1}\sigma/2}{2}\|z_{k+1} - u\|^2\Big) \ . \qquad \square$$

## 4.2 Strongly Convex Theorem

---
**Algorithm 2** KatyushaX$^{\text{s}}$$(F, x_0, b, \eta, \tau, K)$
---

**Input:** function $F(x) = \psi(x) + \frac{1}{n}\sum_{i=1}^n f_i(x)$, starting vector $x_0$, mini-batch size $b \in [n]$, learning rate $\eta > 0$, momentum weight $\tau \in (0, 1]$, $K$ number of epochs;
**Output:** vector $x^{\text{out}}$.

1: $y_{-1} = y_0 \leftarrow x_0$;
2: **for** $k \leftarrow 0$ **to** $K-1$ **do**
3: $\quad x_{k+1} \leftarrow \frac{\frac{3}{2}y_k + \frac{1}{2}x_k - (1-\tau)y_{k-1}}{1+\tau}$;
4: $\quad y_{k+1} \leftarrow \text{SVRG}^{\text{1ep}}(F, x_{k+1}, b, \eta)$;
5: **end for** $\qquad\qquad\qquad\qquad\qquad\qquad\qquad\qquad\qquad\diamond$ *if $\tau = \frac{1}{2}$ then KatyushaX$^{\text{s}}$ = SVRG*
6: **return** $y_K$. $\qquad\qquad\qquad\qquad\qquad\qquad\qquad\qquad\qquad\qquad\diamond$ *smaller $\tau > 0$ means larger momentum*

---

We formally state KatyushaX$^{\text{s}}$ in Algorithm 2 and observe:

**Fact 4.2.** *The updates of KatyushaX$^{\text{s}}$ can be equivalently written as follows. Starting from $z_0 = y_0 = x_0$. Then, in each iteration $k = 0, 1, \ldots, K-1$,*

- $x_{k+1} \leftarrow \tau z_k + (1-\tau)y_k$;
- $y_{k+1} \leftarrow \text{SVRG}^{\text{1ep}}(F, x_{k+1}, b, \eta)$;
- $z_{k+1} \leftarrow \arg\min_{z\in\mathbb{R}^d}\big\{\frac{1}{2}\|z - z_k\|^2 + \langle \frac{x_{k+1} - y_{k+1}}{2\tau}, z\rangle + \frac{\tau}{2}\|z - y_{k+1}\|^2\big\} = \frac{z_k + \tau y_{k+1} - \frac{1}{2\tau}(x_{k+1} - y_{k+1})}{1+\tau}$.

*Proof.* Can be seen by eliminating the sequence $\{z_k\}_k$ from the recurrence. $\qquad\square$

If we choose $\tau_k = \tau \stackrel{\text{def}}{=} \frac{\sqrt{m\eta\sigma}}{2}$ and $\alpha_{k+1} = \frac{m\eta}{2\tau} = \frac{2\tau}{\sigma}$ for all $k \geq 0$, then Fact 4.2 tells us KatyushaX$^{\text{s}}$ falls into General Framework (4.1). One can thus telescope Lemma 4.1 carefully to obtain:



**Theorem 4.3.** *Let $b \in [n]$ be the mini-batch size and $m = \min\{\lceil \frac{n}{b} \rceil, 2\}$ be the epoch length of $\texttt{SVRG}^{\textsf{1ep}}$. If $f(\cdot)$ is $\sigma_f$-strongly convex and $\psi(\cdot)$ is $\sigma_\psi$-strongly convex for some $\sigma \stackrel{\text{def}}{=} \sigma_f + \sigma_\psi > 0$, then selecting $\eta = \min\left\{\frac{1}{2L}, \frac{1}{2\sqrt{\ell_1 \ell_2 m/b}}\right\}$ and $\tau = \min\left\{\frac{1}{2}, \frac{\sqrt{m\eta\sigma}}{2}\right\} \in (0, \frac{1}{2}]$ in $\texttt{KatyushaX}^{\textsf{s}}$, we have*

$$F(y_K) - F(x^*) \leq \frac{2}{(1+\tau)^K}\big(F(y_0) - F(x^*)\big) \ .$$

*Proof.* We only prove the case when $m\eta\sigma \leq 1$. If $m\eta\sigma > 1$, then we can replace $\sigma$ (for analysis purpose only) with a smaller value $\frac{1}{m\eta}$ and the rest of the proof still holds.

Let us choose $\tau_k = \tau \stackrel{\text{def}}{=} \frac{\sqrt{m\eta\sigma}}{2}$ and $\alpha_{k+1} = \frac{m\eta}{2\tau} = \frac{2\tau}{\sigma}$ for all $k \geq 0$. Now, Fact 4.2 tells us that $\texttt{KatyushaX}^{\textsf{s}}$ falls into General Framework (4.1) with $\alpha_{k+1}$ and $\tau_k$. Using Lemma 4.1 with $u = x^*$ we have

$$\frac{2}{\sigma}\big(F(y_{k+1}) - F(x^*)\big) \leq \frac{2(1-\tau)}{\sigma}\big(F(y_k) - F(x^*)\big) + \left(\frac{1}{2}\|z_k - x^*\|^2 - \frac{1+\tau}{2}\|z_{k+1} - x^*\|^2\right) \ .$$

Using $1 - \tau \leq \frac{1}{1+\tau}$, we can telescope the above inequality and get

$$F(y_K) - F(x^*) \leq \frac{1}{(1+\tau)^K}\left(F(y_0) - F(x^*) + \frac{(1+\tau)\sigma}{4}\|z_0 - x^*\|^2\right) \leq \frac{2}{(1+\tau)^K}\big(F(y_0) - F(x^*)\big)$$

where the last inequality uses $x_0 = y_0 = z_0$ and the strong convexity of $F(\cdot)$. □

**Corollary 4.4.** *If $\sigma > 0$, $\texttt{KatyushaX}^{\textsf{s}}$ finds a point $x^{\textsf{out}}$ satisfying $\mathbb{E}[F(x^{\textsf{out}}) - F(x^*)] \leq \varepsilon$ in*

$$T_{\textsf{grad}} = O\left(n + \frac{\sqrt{Lbn}}{\sqrt{\sigma}} + \frac{(\ell_1 \ell_2)^{1/4} n^{3/4}}{\sqrt{\sigma}}\right) \cdot \log \frac{F(x_0) - F(x^*)}{\varepsilon}$$

*computations of stochastic gradients (in expectation), or equivalently*

$$T_{\textsf{iter}} = O\left(\frac{n}{b} + \frac{\sqrt{Ln}}{\sqrt{\sigma b}} + \frac{(\ell_1 \ell_2)^{1/4} n^{3/4}}{\sqrt{\sigma b}}\right) \cdot \log \frac{F(x_0) - F(x^*)}{\varepsilon}$$

*parallel iterations (in expectation).*

In the case $b = 1$ and $\ell_1 = \ell_2 = L$, this implies Theorem 1.

### 4.3 Weakly Convex Case

---
**Algorithm 3** $\texttt{KatyushaX}^{\textsf{w}}(F, x_0, b, \eta, K)$

---
**Input:** function $F(x) = \psi(x) + \frac{1}{n}\sum_{i=1}^n f_i(x)$, starting vector $x_0$, mini-batch size $b \in [n]$, learning rate $\eta > 0$, $K$ number of epochs;
**Output:** vector $x^{\textsf{out}}$.
1: $y_{-1} = y_0 \leftarrow x_0$;
2: **for** $k \leftarrow 0$ **to** $K - 1$ **do**
3:     $x_{k+1} \leftarrow \frac{(3k+1)y_k + (k+1)x_k - (2k-2)y_{k-1}}{2k+4}$;
4:     $y_{k+1} \leftarrow \texttt{SVRG}^{\textsf{1ep}}(F, x_{k+1}, b, \eta)$;
5: **end for**                                                                                                             ⋄ *momentum is auto-computed*
6: **return** $y_K$.

---

We state $\texttt{KatyushaX}^{\textsf{w}}$ in Algorithm 3 and observe:



**Fact 4.5.** *The updates of* `KatyushaX`[w] *can be equivalently written as follows. Starting from $z_0 = y_0 = x_0$. Then, in each iteration $k = 0, 1, \ldots, K-1$,*

- $x_{k+1} \leftarrow \tau_k z_k + (1-\tau_k) y_k$ *for* $\tau_k \stackrel{\text{def}}{=} \frac{2}{k+2}$;
- $y_{k+1} \leftarrow \text{SVRG}^{\text{1ep}}(F, x_{k+1}, b, \eta)$;
- $z_{k+1} \leftarrow z_k - \frac{1}{2\tau_k}(x_{k+1} - y_{k+1}) = \arg\min_{z \in \mathbb{R}^d} \left\{ \frac{1}{2}\|z - z_k\|^2 + \left\langle \frac{x_{k+1} - y_{k+1}}{2\tau_k}, z \right\rangle \right\}$.

*Proof.* Can be seen by eliminating the sequence $\{z_k\}_k$ from the recurrence. □

If we choose $\alpha_{k+1} = \frac{m\eta}{2\tau_k}$ for all $k \geq 0$ and $\sigma = 0$, then Fact 4.2 tells us `KatyushaX`[w] falls into General Framework (4.1). One can thus telescope Lemma 4.1 carefully to obtain:

**Theorem 4.6.** *Let $b \in [n]$ be the mini-batch size and $m = \min\{\lceil \frac{n}{b} \rceil, 2\}$ be the epoch length of* `SVRG`[1ep]. *Selecting $\eta = \min\left\{\frac{1}{2L}, \frac{1}{2\sqrt{\ell_1 \ell_2 m/b}}\right\}$ in* `KatyushaX`[w], *we have*

$$\mathbb{E}[F(y_K) - F(x^*)] \leq \frac{4}{(K+1)^2 m\eta} \|x_0 - x^*\|^2 \ .$$

*Proof.* Let us choose $\tau_k = \frac{2}{k+2}$ and $\alpha_{k+1} = \frac{m\eta}{2\tau_k}$ for all $k \geq 0$. Now, Lemma 4.1 with $u = x^*$ and $\sigma = 0$ becomes

$$\frac{m\eta}{2\tau_k^2}\left(F(y_{k+1}) - F(x^*)\right) \leq \frac{(1-\tau_k) m\eta}{2\tau_k^2}\left(F(y_k) - F(x^*)\right) + \left(\frac{1}{2}\|z_k - x^*\|^2 - \frac{1}{2}\|z_{k+1} - x^*\|^2\right) \ .$$

Using the fact that $\frac{1-\tau_k}{\tau_k^2} < \frac{1}{\tau_{k-1}^2}$ and $\tau_0 = 1$, we can telescope the above inequality and obtain

$$\frac{m\eta}{2\tau_{K-1}^2}\left(F(y_K) - F(x^*)\right) \leq \frac{1}{2}\|z_0 - x^*\|^2 \ . \qquad \square$$

**Corollary 4.7.** `KatyushaX`[w] *finds a point $x^{\text{out}}$ satisfying $\mathbb{E}[F(x^{\text{out}}) - F(x^*)] \leq \varepsilon$ in*

$$T_{\text{grad}} = O\left(n + \frac{\|x_0 - x^*\|}{\sqrt{\varepsilon}} \cdot \left(\sqrt{Lbn} + (\ell_1 \ell_2)^{1/4} n^{3/4}\right)\right)$$

*computations of stochastic gradients (in expectation), or equivalently*

$$T_{\text{iter}} = O\left(\frac{n}{b} + \frac{\|x_0 - x^*\|}{\sqrt{\varepsilon}} \cdot \left(\frac{\sqrt{Ln}}{\sqrt{b}} + \frac{(\ell_1 \ell_2)^{1/4} n^{3/4}}{b}\right)\right)$$

*parallel iterations (in expectation).*

In the case $b = 1$ and $\ell_1 = \ell_2 = L$, this implies Theorem 2.

## 5 KatyushaX Using Non-Uniform Sampling

In this section, we consider a simple variant of Problem (2.1). Instead of assuming all functions $f_i(x)$ have the same $(\ell_1, \ell_2)$-smoothness, we want to assume each $f_i(x)$ is $L_i$-smooth for some possibly different value $L_i > 0$. In practical problems (such as PCA) such parameters $L_i$ may be easily computable.[10] In this setting, the same `KatyushaX`[s] and `KatyushaX`[w] methods still apply but their convergence rates will scale in $\sqrt{\max_i L_i}$.

We improve such rates by introducing non-uniform sampling. This requires only one simple change to `SVRG`[1ep] which redefines the gradient estimator $\widetilde{\nabla}_t$ as follows:

---

[10] In principle, one can more generally assume each $f_i(x)$ is $(\ell_{1,i}, \ell_{2,i})$-smooth. Unfortunately, we do not have a clean complexity statement for this case, so have simply assumed $\ell_{1,i} = \ell_{2,i} = L_i$.



- Let distribution $\mathcal{D}$ be to select $i \in [n]$ with probability $p_i \stackrel{\text{def}}{=} \frac{L_i^2}{\sum_j L_j^2}$;
- Let $S_t$ be $b$ i.i.d. samples from $\mathcal{D}$ with replacement;
- $\widetilde{\nabla}_t \leftarrow \frac{1}{b} \sum_{i \in S_t} \frac{1}{np_i} (\nabla f_i(w_t) - \nabla f_i(w_0))$.

Denoting by $\overline{L} \stackrel{\text{def}}{=} \left( \sum_i L_i^2 / n \right)^{1/2}$, it is easy to derive that $\overline{L} \geq \frac{L_1 + \cdots + L_n}{n} \geq L$ where $L$ is the smoothness of $L$.

Using this new gradient estimator $\widetilde{\nabla}_t$, we have the following new variance upper bound (proved in Appendix C).

**Lemma 5.1.** $\mathbb{E}_{S_t}[\widetilde{\nabla}_t] = \nabla f(w_t)$ and $\mathbb{E}\big[\|\widetilde{\nabla}_t - \nabla f(w_t)\|^2\big] \leq \frac{\overline{L}^2}{b} \|w_t - w_0\|^2$ .

Therefore, one can replace $Q = \frac{\ell_1 \ell_2}{b}$ as used for Lemma 3.3 with this new choice $Q = \frac{\overline{L}^2}{b}$. The exact same proof shows that `KatyushaX`$^{\text{s}}$ now finds a point $x^{\text{out}}$ satisfying $\mathbb{E}[F(x^{\text{out}}) - F(x^*)] \leq \varepsilon$ in

$$T_{\text{grad}} = O\left( n + \frac{\sqrt{Lbn}}{\sqrt{\sigma}} + \frac{\overline{L}^{1/2} n^{3/4}}{\sqrt{\sigma}} \right) \cdot \log \frac{F(x_0) - F(x^*)}{\varepsilon}$$

computations of stochastic gradients (in expectation), and `KatyushaX`$^{\text{w}}$ now finds a point $x^{\text{out}}$ satisfying $\mathbb{E}[F(x^{\text{out}}) - F(x^*)] \leq \varepsilon$ in

$$T_{\text{grad}} = O\left( n + \frac{\|x_0 - x^*\|}{\sqrt{\varepsilon}} \cdot \left( \sqrt{Lbn} + \overline{L}^{1/2} n^{3/4} \right) \right)$$

computations of stochastic gradients (in expectation).

# Appendix

## A Proofs for Section 2

**Fact 2.3.** *Given any sequence $D_0, D_1, \ldots$ of reals, if $N \sim \text{Geom}(p)$, then*
- $\mathbb{E}_N[D_N - D_{N+1}] = \frac{p}{1-p} \mathbb{E}[D_0 - D_N]$, and
- $\mathbb{E}_N[D_N] = (1-p) \mathbb{E}[D_{N+1}] + p D_0$

*Proof.* We only prove the first item since the second is by rearranging the first.

$$\mathbb{E}[D_N - D_{N+1}] = \sum_{k \geq 0} (D_k - D_{k+1})(1-p)^k p = p D_0 + \sum_{k \geq 1} D_k \big( (1-p)^k p - (1-p)^{k-1} p \big)$$

$$= p D_0 - \frac{p}{1-p} \sum_{k \geq 1} D_k (1-p)^k p = \frac{p}{1-p} D_0 - \frac{p}{1-p} \sum_{k \geq 0} D_k (1-p)^k p = \frac{p}{1-p} \mathbb{E}[D_0 - D_N] \ . \quad \square$$

**Lemma 2.4.** *If $h(\cdot)$ is proper convex and $\sigma$-strongly convex and $z_{k+1} = \arg\min_{z \in \mathbb{R}^d} \{ \frac{1}{2\alpha} \|z - z_k\|^2 + \langle \xi, z \rangle + h(z) \}$, then for every $u \in \mathbb{R}^d$, we have*

$$\langle \xi, z_k - u \rangle + h(z_{k+1}) - h(u) \leq \langle \xi, z_k - z_{k+1} \rangle + \frac{\|u - z_k\|^2}{2\alpha} - \frac{(1+\sigma\alpha)\|u - z_{k+1}\|^2}{2\alpha} - \frac{\|z_k - z_{k+1}\|^2}{2\alpha}$$

$$\leq \frac{\alpha}{2} \|\xi\|^2 + \frac{\|u - z_k\|^2}{2\alpha} - \frac{(1+\sigma\alpha)\|u - z_{k+1}\|^2}{2\alpha} \ .$$



*Proof.* Recall that the minimality of $z_{k+1}$ implies the existence of some subgradient $g \in \partial h(z_{k+1})$ which satisfies $\frac{1}{\alpha}(z_{k+1} - z_k) + \xi + g = 0$. Combining this with $h(u) - h(z_{k+1}) \geq \langle g, u - z_{k+1} \rangle + \frac{\sigma}{2}\|u - z_{k+1}\|^2$, which is due to the strong convexity of $h(\cdot)$, we immediately have

$$h(u) - h(z_{k+1}) + \langle \frac{1}{\alpha}(u - z_{k+1}) + \xi, u - z_{k+1} \rangle - \frac{\sigma}{2}\|u - z_{k+1}\|^2 \geq \langle \frac{1}{\alpha}(z_{k+1} - z_k) + \xi + g, u - z_{k+1} \rangle \geq 0.$$

Therefore,

$$\langle \xi, z_k - u \rangle + h(z_{k+1}) - h(u) = \langle \xi, z_k - z_{k+1} \rangle + \langle \xi, z_{k+1} - u \rangle + h(z_{k+1}) - h(u)$$

$$\overset{①}{\leq} \langle \xi, z_k - z_{k+1} \rangle + \langle -\frac{1}{\alpha}(z_{k+1} - z_k), z_{k+1} - u \rangle - \frac{\sigma}{2}\|u - z_{k+1}\|^2$$

$$\overset{②}{=} \langle \xi, z_k - z_{k+1} \rangle + \frac{\|u - z_k\|^2}{2\alpha} - \frac{(1+\sigma\alpha)\|u - z_{k+1}\|^2}{2\alpha} - \frac{\|z_{k+1} - z_k\|^2}{2\alpha}$$

$$\overset{③}{\leq} \frac{\alpha}{2}\|\xi\|^2 + \frac{\|u - z_k\|^2}{2\alpha} - \frac{(1+\sigma\alpha)\|u - z_{k+1}\|^2}{2\alpha}.$$

Above, inequality ① holds for the reasons explained above; equality ② can be verified by expanding out the terms; and inequality ③ is due to Young's inequality. □

## B Proofs for Section 3

**Lemma 3.2.** *If $w_{t+1} = \arg\min_{y \in \mathbb{R}^d} \{\frac{1}{2\eta}\|y - w_t\|^2 + \psi(y) + \langle \widetilde{\nabla}_t, y \rangle\}$ for some random vector $\widetilde{\nabla}_t \in \mathbb{R}^d$ satisfying $\mathbb{E}[\widetilde{\nabla}_t] = \nabla f(w_t)$, then for every $u \in \mathbb{R}^d$, we have*

$$\mathbb{E}\big[F(w_{t+1}) - F(u)\big] \leq \mathbb{E}\Big[\frac{\eta}{2(1-\eta L)}\|\widetilde{\nabla}_t - \nabla f(w_t)\|^2 + \frac{(1-\sigma_f\eta)\|u - w_t\|^2 - (1+\sigma_\psi\eta)\|u - w_{t+1}\|^2}{2\eta}\Big].$$

*Proof.* We first upper bound the left hand side:

$$\mathbb{E}\big[F(w_{t+1}) - F(u)\big] = \mathbb{E}\big[f(w_{t+1}) - f(u) + \psi(w_{t+1}) - \psi(u)\big]$$

$$\overset{①}{\leq} \mathbb{E}\big[f(w_t) + \langle \nabla f(w_t), w_{t+1} - w_t \rangle + \tfrac{L}{2}\|w_t - w_{t+1}\|^2 - f(u) + \psi(w_{t+1}) - \psi(u)\big]$$

$$\overset{②}{\leq} \mathbb{E}\big[\langle \nabla f(w_t), w_t - u \rangle - \tfrac{\sigma_f}{2}\|u - w_t\|^2 + \langle \nabla f(w_t), w_{t+1} - w_t \rangle + \tfrac{L}{2}\|w_t - w_{t+1}\|^2 + \psi(w_{t+1}) - \psi(u)\big]$$

$$\overset{③}{=} \mathbb{E}\big[\langle \widetilde{\nabla}_t, w_t - u \rangle - \tfrac{\sigma_f}{2}\|u - w_t\|^2 + \langle \nabla f(w_t), w_{t+1} - w_t \rangle + \tfrac{L}{2}\|w_t - w_{t+1}\|^2 + \psi(w_{t+1}) - \psi(u)\big].$$

Above, inequalities ① and ② are respectively due to the smoothness and strong-convexity of $f(\cdot)$, and ③ is because $\mathbb{E}[\widetilde{\nabla}_t] = \nabla f(w_t)$. Next, using Lemma 2.4 (with $z_k = w_t$, $z_{k+1} = w_{t+1}$, $h(\cdot) = \psi(\cdot)$, $\xi = \widetilde{\nabla}_t$) we have

$$\langle \widetilde{\nabla}_t, w_t - u \rangle + \psi(w_{t+1}) - \psi(u) \leq \langle \widetilde{\nabla}_t, w_t - w_{t+1} \rangle + \frac{\|u - w_t\|^2}{2\eta} - \frac{(1+\sigma_\psi\eta)\|u - w_{t+1}\|^2}{2\eta} - \frac{\|w_{t+1} - w_t\|^2}{2\eta}.$$

Combining the above two inequalities, we have

$$\mathbb{E}\big[F(w_{t+1}) - F(u)\big]$$

$$\leq \mathbb{E}\Big[\langle \widetilde{\nabla}_t - \nabla f(w_t), w_t - w_{t+1} \rangle - \frac{1-\eta L}{2\eta}\|w_t - w_{t+1}\|^2 + \frac{(1-\sigma_f\eta)\|u - w_t\|^2 - (1+\sigma_\psi\eta)\|u - w_{t+1}\|^2}{2\eta}\Big]$$

$$\overset{④}{\leq} \mathbb{E}\Big[\frac{\eta}{2(1-\eta L)}\|\widetilde{\nabla}_t - \nabla f(w_t)\|^2 + \frac{(1-\sigma_f\eta)\|u - w_t\|^2 - (1+\sigma_\psi\eta)\|u - w_{t+1}\|^2}{2\eta}\Big].$$

Above, ④ is by Young's inequality. □



**Lemma 3.3.** *If the gradient estimator $\widetilde{\nabla}_t$ satisfies $\mathbb{E}[\widetilde{\nabla}_t] = \nabla f(w_t)$ and $\mathbb{E}\big[\|\widetilde{\nabla}_t - \nabla f(w_t)\|^2\big] \leq Q\|w_0 - w_t\|^2$ for every $t$ for some universal constant $Q > 0$, then as long as $\eta \leq \min\big\{\frac{1}{2L}, \frac{1}{2\sqrt{Qm}}\big\}$, it satisfies*

$$\forall u \in \mathbb{R}^d: \quad \mathbb{E}\big[F(w^+) - F(u)\big] \leq \mathbb{E}\Big[-\frac{1}{4m\eta}\|w^+ - w_0\|^2 + \frac{\langle w_0 - w^+, w_0 - u\rangle}{m\eta} - \frac{\sigma_f + \sigma_\psi}{4}\|w^+ - u\|^2\Big] \ .$$

*Proof.* Since $\eta \leq \frac{1}{2L}$, we can write Lemma 3.2 as

$$\mathbb{E}\big[F(w_{t+1}) - F(u)\big]$$
$$\leq \mathbb{E}\Big[\eta\|\widetilde{\nabla}_t - \nabla f(w_t)\|^2 + \frac{\|u - w_t\|^2 - \|u - w_{t+1}\|^2}{2\eta} - \frac{\sigma_f}{2}\|u - w_t\|^2 - \frac{\sigma_\psi}{2}\|u - w_{t+1}\|^2\Big]$$
$$\leq \mathbb{E}\Big[\eta Q\|w_t - w_0\|^2 + \frac{\|u - w_t\|^2 - \|u - w_{t+1}\|^2}{2\eta} - \frac{\sigma_f}{2}\|u - w_t\|^2 - \frac{\sigma_\psi}{2}\|u - w_{t+1}\|^2\Big] \ . \quad \text{(B.1)}$$

Taking $t = M$ where $M \sim \mathsf{Geom}(\frac{1}{m})$ follows from the geometric distribution, we have

$$\mathbb{E}\big[F(w_{M+1}) - F(u)\big]$$
$$\overset{①}{\leq} \mathbb{E}\Big[\eta Q\|w_M - w_0\|^2 + \frac{\|u - w_0\|^2 - \|u - w_M\|^2}{2(m-1)\eta} - \frac{\sigma_f}{2}\|u - w_M\|^2 - \frac{\sigma_\psi}{2}\|u - w_{M+1}\|^2\Big]$$
$$\overset{②}{=} \mathbb{E}\Big[\frac{m-1}{m}\eta Q\|w_{M+1} - w_0\|^2 + \frac{\|u - w_0\|^2 - \|u - w_{M+1}\|^2}{2m\eta}$$
$$\qquad - \frac{\sigma_f}{2m}\|u - w_0\|^2 - \frac{\sigma_f(m-1)}{2m}\|u - w_{M+1}\|^2 - \frac{\sigma_\psi}{2}\|u - w_{M+1}\|^2\Big]$$
$$\overset{③}{\leq} \mathbb{E}\Big[\eta Q\|w_{M+1} - w_0\|^2 + \frac{\|u - w_0\|^2 - \|u - w_{M+1}\|^2}{2m\eta} - \frac{\sigma_f + \sigma_\psi}{4}\|u - w_{M+1}\|^2\Big]$$
$$\overset{④}{\leq} \mathbb{E}\Big[-\frac{1}{4m\eta}\|w_0 - w_{M+1}\|^2 + \frac{\|u - w_0\|^2 - \|u - w_{M+1}\|^2 + \|w_0 - w_{M+1}\|^2}{2m\eta} - \frac{\sigma_f + \sigma_\psi}{4}\|w_{M+1} - u\|^2\Big]$$
$$\overset{⑤}{=} \mathbb{E}\Big[-\frac{1}{4m\eta}\|w_{M+1} - w_0\|^2 + \frac{\langle w_0 - w_{M+1}, w_0 - u\rangle}{m\eta} - \frac{\sigma_f + \sigma_\psi}{4}\|w_{M+1} - u\|^2\Big]$$

Above, inequality ① is by substituting $t = M$ into (B.1) and applying Fact 2.3; equality ② is by applying Fact 2.3; inequality ③ uses $m \geq 2$; inequality ④ uses our assumption $\eta \leq \frac{1}{2\sqrt{Qm}}$; and equality ⑤ can be verified by expanding out the norms. □

## C Variance Bonds for Gradient Estimators

In this section, we provide two types of variance upper bound of the form $\mathbb{E}\big[\|\widetilde{\nabla}_t - \nabla f(w_t)\|^2\big] \leq Q\|w_t - w_0\|^2$. The first type is for uniform sampling over the stochastic gradients, and the second type is for non-uniform sampling. Both of them are not hard to prove.

### C.1 Uniform Sampling

**Lemma 3.4.** *If each $f_i(x)$ is $(\ell_1, \ell_2)$-smooth for $\ell_2 \geq \ell_1$, and the gradient estimator*

$$\widetilde{\nabla}_t = \mu + \frac{1}{b}\sum_{i \in S_t}\big(\nabla f_i(w_t) - \nabla f_i(w_0)\big)$$

*where $S_t$ consists of $b$ i.i.d. uniform random indices from $[n]$ with replacement. Then, $\mathbb{E}_{S_t}[\widetilde{\nabla}_t] = \nabla f(w_t)$ and $\mathbb{E}_{S_t}\big[\|\widetilde{\nabla}_t - \nabla f(w_t)\|^2\big] \leq \frac{\ell_1 \ell_2}{b}\|w_t - w_0\|^2$ .*



*Proof.* We calculate that

$$\mathbb{E}\big[\|\widetilde{\nabla}_t - \nabla f(w_t)\|^2\big] = \mathbb{E}_{S_t}\Big[\Big\|\Big(\frac{1}{b}\sum_{i\in S_t}\big(\nabla f(w_0) + \nabla f_i(w_t) - \nabla f_i(w_0)\big)\Big) - \nabla f(w_t)\Big\|^2\Big]$$

$$\stackrel{\text{①}}{=} \frac{1}{b}\mathbb{E}_{i\in_R[n]}\Big[\Big\|\big(\nabla f(w_0) + \nabla f_i(w_t) - \nabla f_i(w_0)\big) - \nabla f(w_t)\Big\|^2\Big]$$

$$= \frac{1}{b}\mathbb{E}_{i\in_R[n]}\Big[\Big\|\big(\nabla f_i(w_t) - \nabla f_i(w_0)\big) - \big(\nabla f(w_t) - f(w_0)\big)\Big\|^2\Big]$$

$$\stackrel{\text{②}}{\leq} \frac{1}{b}\mathbb{E}_{i\in_R[n]}\Big[\Big\|\big(\nabla f_i(w_t) - \nabla f_i(w_0)\big)\Big\|^2\Big]$$

$$\stackrel{\text{③}}{\leq} \frac{1}{b}\Big((\ell_1 - \ell_2)\langle\nabla f(w_t) - \nabla f(w_0), w_t - w_0\rangle + \ell_1\ell_2\|w_t - w_0\|^2\Big)$$

$$\stackrel{\text{④}}{\leq} \frac{\ell_1\ell_2}{b}\|w_t - w_0\|^2 \ .$$

Above, equality ① follows because for i.i.d. random vectors $\zeta_1, \ldots, \zeta_b$, we have $\mathbb{E}\big[\|\frac{1}{b}\sum_{j\in[b]}\zeta_j - \mathbb{E}[\zeta_j]\|^2\big] = \frac{1}{b}\mathbb{E}[\|\zeta_1 - \mathbb{E}[\zeta_1]\|^2]$; inequality ② is because for any random vector $\zeta \in \mathbb{R}^d$, it holds that $\mathbb{E}\|\zeta - \mathbb{E}\zeta\|^2 = \mathbb{E}\|\zeta\|^2 - \|\mathbb{E}\zeta\|^2$; inequality ③ is due to Claim C.1 below and $\mathbb{E}_{i\in_R[n]}[\nabla f_i(x)] = \nabla f(x)$; and inequality ④ follows because $\ell_1 \leq \ell_2$ and for any convex function $g(\cdot)$, we have $\langle\nabla g(x) - \nabla g(y), x - y\rangle \geq (g(x) - g(y)) + (g(y) - g(x)) = 0$. □

**Claim C.1.** *If $f_i(\cdot)$ is an $(\ell_1, \ell_2)$-smooth function, then*

$$\|\nabla f_i(x) - \nabla f_i(y)\|^2 \leq (\ell_1 - \ell_2)\langle\nabla f_i(x) - \nabla f_i(y), x - y\rangle + \ell_1\ell_2\|y - x\|^2 \ .$$

*Proof.* Define $\phi_i(z) \stackrel{\text{def}}{=} -f_i(z) + \langle\nabla f_i(x), z\rangle + \frac{\ell_1}{2}\|z - x\|^2$ for each $i \in [n]$. It is clear that $\phi_i(y)$ is a convex, $(\ell_1 + \ell_2)$-smooth function that has a minimizer $z = x$ (which can be seen by taking the derivative). For this reason, we claim that

$$\phi_i(x) \leq \phi_i\Big(z - \frac{\nabla\phi_i(z)}{\ell_1 + \ell_2}\Big) \leq \phi_i(z) - \frac{1}{2(\ell_1 + \ell_2)}\|\nabla\phi_i(z)\|^2 \ , \tag{C.1}$$

for each $z$, and this inequality is classical for smooth functions (see for instance Theorem 2.1.5 in the textbook [22]). By expanding out the definition of $\phi_i(\cdot)$ and choosing $z = y$ in (C.1), we immediately have

$$-f_i(x) + \langle\nabla f_i(x), x\rangle \leq -f_i(y) + \langle\nabla f_i(x), y\rangle + \frac{\ell_1}{2}\|y - x\|^2$$
$$- \frac{1}{2(\ell_1 + \ell_2)}\|\nabla f_i(y) - \nabla f_i(x) - \ell_1(y - x)\|^2 \ .$$

Summing up the above inequality with itself but swapping $x$ and $y$, we have

$$\langle\nabla f_i(x) - \nabla f_i(y), x - y\rangle \leq \ell_1\|y - x\|^2 - \frac{1}{\ell_1 + \ell_2}\|\nabla f_i(y) - \nabla f_i(x) - \ell_1(y - x)\|^2 \ .$$

Or equivalently

$$\frac{\ell_2 - \ell_1}{\ell_1 + \ell_2}\langle\nabla f_i(x) - \nabla f_i(y), x - y\rangle \leq \ell_1\|y - x\|^2 - \frac{1}{\ell_1 + \ell_2}\|\nabla f_i(y) - \nabla f_i(x)\|^2 - \frac{\ell_1^2}{\ell_1 + \ell_2}\|y - x\|^2 \ .$$

□



## C.2 Non-Uniform Sampling

**Lemma 5.1.** *Suppose each $f_i(x)$ is $L_i$-smooth for some possibly different parameter $L_i$, and let $\overline{L} \stackrel{\text{def}}{=} \big(\sum_i L_i^2/n\big)^{1/2}$. Then, we can define $\widetilde{\nabla}_t$ as follows:*

- *Let distribution $\mathcal{D}$ be to select $i \in [n]$ with probability $p_i \stackrel{\text{def}}{=} \frac{L_i^2}{\sum_j L_j^2}$;*
- *Let $S_t$ be $b$ i.i.d. samples from $\mathcal{D}$ with replacement;*
- *$\widetilde{\nabla}_t \leftarrow \frac{1}{b}\sum_{i\in S_t} \frac{1}{np_i}\big(\nabla f_i(w_t) - \nabla f_i(w_0)\big).$*

*Then, $\mathbb{E}_{S_t}[\widetilde{\nabla}_t] = \nabla f(w_t)$ and $\mathbb{E}\big[\|\widetilde{\nabla}_t - \nabla f(w_t)\|^2\big] \leq \frac{\overline{L}^2}{b}\|w_t - w_0\|^2$ .*

*Proof.*

$$\mathbb{E}\big[\|\widetilde{\nabla}_t - \nabla f(w_t)\|^2\big] = \mathbb{E}_{S_t}\bigg[\bigg\|\Big(\frac{1}{b}\sum_{i\in S_t}\Big(\nabla f(w_0) + \frac{1}{np_i}\big(\nabla f_i(w_t) - \nabla f_i(w_0)\big)\Big)\Big) - \nabla f(w_t)\bigg\|^2\bigg]$$

$$\stackrel{①}{=} \frac{1}{b}\mathbb{E}_{i\sim\mathcal{D}}\bigg[\bigg\|\Big(\nabla f(w_0) + \frac{1}{np_i}\big(\nabla f_i(w_t) - \nabla f_i(w_0)\big)\Big) - \nabla f(w_t)\bigg\|^2\bigg]$$

$$= \frac{1}{b}\mathbb{E}_{i\sim\mathcal{D}}\bigg[\bigg\|\frac{1}{np_i}\big(\nabla f_i(w_t) - \nabla f_i(w_0)\big) - \big(\nabla f(w_t) - f(w_0)\big)\bigg\|^2\bigg]$$

$$\stackrel{②}{\leq} \frac{1}{b}\mathbb{E}_{i\sim\mathcal{D}}\bigg[\bigg\|\frac{1}{np_i}\big(\nabla f_i(w_t) - \nabla f_i(w_0)\big)\bigg\|^2\bigg]$$

$$\stackrel{②}{\leq} \frac{1}{b} \cdot \sum_{i\in[n]} \frac{L_i^2}{n^2 p_i}\|w_t - w_0\|^2 = \frac{\overline{L}^2}{b}\|w_t - w_0\|^2 \enspace.$$

Above, equality ① follows because for i.i.d. random vectors $\zeta_1,\dots,\zeta_b$, we have $\mathbb{E}\big[\big\|\frac{1}{b}\sum_{j\in[b]}\zeta_j - \mathbb{E}[\zeta_j]\big\|^2\big] = \frac{1}{b}\mathbb{E}[\|\zeta_1 - \mathbb{E}[\zeta_1]\|^2]$; inequality ② is because for any random vector $\zeta \in \mathbb{R}^d$, it holds that $\mathbb{E}\|\zeta - \mathbb{E}\zeta\|^2 = \mathbb{E}\|\zeta\|^2 - \|\mathbb{E}\zeta\|^2$; and inequality ③ uses the smoothness definition of $f_i(\cdot)$. □

## D Improved SVRG Convergence Rate

In this section we consider a simple SVRG algorithm that, starting from $x_0 \in \mathbb{R}^d$, repeatedly performs updates $x_{k+1} \leftarrow \text{SVRG}^{1\text{ep}}(F, x_k, b, \eta)$. We refer to this algorithm as SVRG in this section.

**Theorem D.1.** *If $\sigma > 0$ and $\eta > 0$ satisfies Theorem 3.1, then setting $\tau = \min\big\{\frac{m\eta\sigma}{2}, \frac{1}{2}\big\}$ and $\overline{x} = \frac{(1+\tau)^K - 1}{\tau}\sum_{k=0}^{K-1}(1+\tau)^k x_{k+1}$, we have*

$$\mathbb{E}[F(\overline{x}) - F(x^*)] \leq O((1+\tau)^{-K})\big(F(x_0) - F(x^*)\big) \enspace.$$

*In other words, choosing $\eta = \min\big\{\frac{1}{2L}, \frac{1}{2\sqrt{\ell_1\ell_2 m/b}}\big\}$, SVRG finds $\mathbb{E}[F(\overline{x}) - F(x^*)] \leq \varepsilon$ in*

$$T_{\text{grad}} = O\Big(\Big(n + \frac{bL + \sqrt{\ell_1\ell_2}n}{\sigma}\Big)\log\frac{F(x_0) - F(x^*)}{\varepsilon}\Big)$$

*stochastic gradient computations, or equivalently*

$$T_{\text{grad}} = O\Big(\Big(\frac{n}{b} + \frac{L + \sqrt{\ell_1\ell_2}n/b}{\sigma}\Big)\log\frac{F(x_0) - F(x^*)}{\varepsilon}\Big)$$

*parallel iterations.*



*Proof of Theorem D.1.* To analyze SVRG, we rewrite Theorem 3.1 equivalently as:

$$\mathbb{E}\big[F(w^+) - F(u)\big] \leq \mathbb{E}\Big[\frac{1}{4m\eta}\|w^+ - w_0\|^2 + \frac{1}{2m\eta}\|w_0 - u\|^2 - \frac{1}{2m\eta}\|w^+ - u\|^2 - \frac{\sigma}{4}\|w^+ - u\|^2\Big] \ . \tag{D.1}$$

We now apply (D.1) twice, once with $w_0 = x_k$, $w^+ = x_{k+1}$, $u = x^*$ and once with $w_0 = x_k$, $w^+ = x_{k+1}$, $u = x_k$. Summing them up, we get

$$\mathbb{E}\big[2F(x_{k+1}) - F(x^*) - F(x_k)\big] \leq \mathbb{E}\Big[\frac{1}{2m\eta}\|x_k - x^*\|^2 - \frac{1}{2m\eta}\|x_{k+1} - x^*\|^2 - \frac{\sigma}{4}\|x_{k+1} - x^*\|^2\Big] \ . \tag{D.2}$$

Let $\tau = \min\big\{\frac{m\eta\sigma}{2}, \frac{1}{2}\big\}$. Multiplying both sides of (D.2) by $(1+\tau)^k$, we obtain

$$\mathbb{E}\big[2(1+\tau)^k(F(x_{k+1}) - F(x^*)) - (1+\tau)^k(F(x_k) - F(x^*))\big]$$
$$\leq \mathbb{E}\Big[\frac{(1+\tau)^k}{2m\eta}\|x_k - x^*\|^2 - \frac{(1+\tau)^{k+1}}{2m\eta}\|x_{k+1} - x^*\|^2\Big] \ .$$

Telescoping it for $k = 0, 1, \ldots, K-1$, we have

$$\mathbb{E}\Big[\sum_{k=0}^{K-1}(1-\tau)(1+\tau)^k(F(x_{k+1}) - F(x^*))\Big] \leq F(x_0) - F(x^*) + \frac{1}{2m\eta}\|x_0 - x^*\|^2$$
$$\leq \big(1 + \frac{1}{m\eta\sigma}\big)\big(F(x_0) - F(x^*)\big)$$
$$= \big(\frac{1}{2\tau} + 1\big)\big(F(x_0) - F(x^*)\big) \ .$$

Using the definition $\bar{x} = \frac{(1+\tau)^K - 1}{\tau}\sum_{k=0}^{K-1}(1+\tau)^k x_{k+1}$ and the convexity of $f(x)$, we immediately conclude that

$$F(\bar{x}) - F(x^*) \leq O((1+\tau)^{-K})\big(F(x_0) - F(x^*)\big) \ . \qquad \square$$

**Theorem D.2.** *In the case of $\sigma = 0$, a variant of* SVRG *achieves the following complexity:*

$$T_{\mathsf{grad}} = O\Big(n\log\frac{F(x_0) - F(x^*)}{\varepsilon} + \frac{(bL + \sqrt{\ell_1\ell_2 n})\|x_0 - x^*\|^2}{\varepsilon}\Big)$$

*stochastic gradient computations, or equivalently*

$$T_{\mathsf{grad}} = O\Big(\frac{n}{b}\log\frac{F(x_0) - F(x^*)}{\varepsilon} + \frac{(L + \sqrt{\ell_1\ell_2 n}/b)\|x_0 - x^*\|^2}{\varepsilon}\Big)$$

*parallel iterations.*

*Proof sketch.* The simplest way to show this result is by applying the optimal reduction [6] to reduce the problem to the strongly convex case (and thus Theorem D.1). One can also design a direct method with such complexity by doubling the epoch length $m$ from epoch to epoch, following a similar analysis from [11]. Since the main focus of this paper is to design *accelerated* method, we ignore the details here. $\qquad\square$